\documentclass[11pt]{article}

\usepackage[T1]{fontenc}
\usepackage[utf8]{inputenc}
\usepackage{lmodern}
\usepackage[defaultlines=2,all]{nowidow}
\usepackage[protrusion=true,expansion=true]{microtype}

\usepackage[a4paper,margin=1in]{geometry}
\usepackage{setspace}
\usepackage{amsmath,amssymb,amsfonts}
\usepackage{bm}

\usepackage{graphicx}
\usepackage[labelfont=bf,labelsep=period]{caption}
\usepackage{subcaption}
\usepackage{booktabs}
\usepackage{multirow}
\usepackage{siunitx}
\usepackage{float} 

\usepackage{xcolor}
\definecolor{linkblue}{RGB}{0,80,170}
\usepackage[
  colorlinks=true,
  linkcolor=linkblue,
  citecolor=linkblue,
  urlcolor=linkblue
]{hyperref}
\usepackage[nameinlink,noabbrev]{cleveref}

\usepackage{url}

\newcommand{\Pleaf}{P^{\text{leaf}}}
\newcommand{\Pens}{P^{\text{ens}}}
\newcommand{\Pboost}{P^{\text{boost}}}
\newcommand{\mtb}{\textit{Mycobacterium tuberculosis}}
\newcommand{\MSE}{\mathrm{MSE}}

\title{\vspace{-1.5em}\textbf{\LARGE Evaluating Double Descent in Machine Learning:}\\[0.25em]
\textbf{\LARGE Insights from Tree-Based Models Applied to a Genomic Prediction Task}}
\author{
  \large Guillermo Comesa\~na Cimadevila\\[0.25em]
  \normalsize Department of Life Sciences, University of Bath\\
  \normalsize \texttt{gcc46@bath.ac.uk}
}
\date{} 

\begin{document}
\maketitle

\vspace{-1.0em} 

\begin{abstract}
Classical learning theory describes a well-characterised U-shaped relationship between model complexity and prediction error, reflecting a transition from underfitting in underparameterised regimes to overfitting as complexity grows. Recent work, however, has introduced the notion of a second descent in test error beyond the interpolation threshold---giving rise to the so-called double descent phenomenon. While double descent has been studied extensively in the context of deep learning, it has also been reported in simpler models, including decision trees and gradient boosting. In this work, we revisit these claims through the lens of classical machine learning applied to a biological classification task: predicting isoniazid resistance in \mtb{} using whole-genome sequencing data. We systematically vary model complexity along two orthogonal axes---learner capacity (e.g., $\Pleaf$, $\Pboost$) and ensemble size (i.e., $\Pens$)---and show that double descent consistently emerges only when complexity is scaled jointly across these axes. When either axis is held fixed, generalisation behaviour reverts to classical U- or L-shaped patterns. These results are replicated on a synthetic benchmark and support the unfolding hypothesis, which attributes double descent to the projection of distinct generalisation regimes onto a single complexity axis. Our findings underscore the importance of treating model complexity as a multidimensional construct when analysing generalisation behaviour. All code and reproducibility materials are available at: \url{https://github.com/guillermocomesanacimadevila/Demystifying-Double-Descent-in-ML}.

\end{abstract}

\vspace{1.25em}
\noindent\rule{\linewidth}{0.4pt}
\vspace{1.25em}

\section{Introduction}
\begin{figure}[H]
  \centering
  \includegraphics[width=0.8\linewidth]{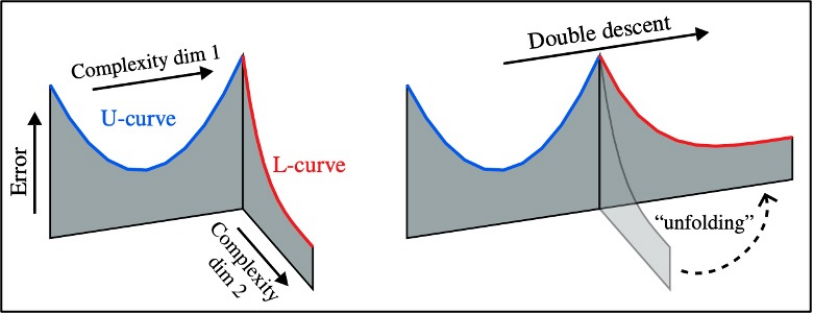}
  \caption{Double descent illustration emerging from two complexity axes. Left: error varies across two model complexity dimensions, forming a U-curve (blue) along one axis and an L-curve (red) along the other. Right: collapsing these dimensions produces the double descent curve, suggesting it may arise from merging distinct generalisation behaviours. Figure adapted from Curth et~al.\ \cite{Curth2023}.}
  \label{fig:concept}
\end{figure}

The traditional relationship between model complexity and prediction error has long been explained by the bias--variance trade-off, which posits that prediction error follows a U-shaped curve (Figure~\ref{fig:concept}, left pannel) as model complexity increases \cite{Domingos2000,James2021}. In this framework, models with insufficient complexity exhibit high bias and underfit the data, while overly complex models tend to memorise the training data, leading to high variance and poor generalisation to unseen inputs (i.e., overfitting) \cite{RajnarayanWolpert2008,James2021}. The optimal predictive performance is thought to lie at an intermediate point of complexity, where bias and variance are minimised \cite{Briscoe2011}. This foundational concept underpins widely used model selection strategies such as cross-validation, regularisation, and information-theoretic criteria, including the Akaike and Bayesian Information Criteria \cite{Fieldsend2008}.

This view implicitly assumes that increasing model complexity beyond the interpolation threshold---where the number of model parameters equals the number of training samples---would continue to degrade generalisation \cite{Geman1992,Vapnik2000}. However, recent empirical findings in modern machine learning challenge this assumption. Notably, overparameterised models such as deep neural networks can achieve near-zero training error and yet continue to generalise effectively, defying the predictions of the classical U-shaped error curve \cite{Goodfellow2016,Belkin2021,Bartlett2020}. To account for this observation, Belkin et~al.\ \cite{Belkin2019} proposed the double descent phenomenon (Figure~\ref{fig:concept}, right pannel), wherein test error initially decreases with complexity, rises near the interpolation threshold, and then decreases again as complexity increases further. The double descent framework suggests that increasing model complexity can, under certain conditions, lead to improved generalisation even in highly overparameterised regimes \cite{Lafon2024}. While originally observed in deep learning models, subsequent work has shown that double descent can emerge in simpler settings, including kernel methods, decision trees, and even ordinary least-squares regression \cite{Christensen2024,Belkin2019}. 

\vspace{1em}

Nevertheless, its underlying theoretical basis remains a topic of ongoing debate \cite{SaCouto2022}. Recent critiques argue that double descent may be a visual artefact of collapsing multidimensional model complexity into a single axis \cite{Schaeffer2023,SaCouto2022}. Curth et~al.\ \cite{Curth2023} built on this view by proposing that the observed curve arises from the projection of two separate generalisation regimes---the classical bias--variance trade-off and a high-dimensional interpolation regime---onto a shared axis. In this formulation, double descent does not reflect a continuous generalisation phenomenon but rather the unfolding of separate complexity dynamics \cite{Curth2023} (Figure~\ref{fig:concept}). Despite growing theoretical interest, empirical studies of double descent remain limited. Prior work has primarily focused on least squares regression or deep learning architectures, with few investigations in classical tree-based models such as decision trees and gradient boosting---Curth et~al.\ \cite{Curth2023} being a notable exception. Moreover, the presence of double descent in real-world biological datasets remains unexplored. In this study, we address this gap by applying the double descent framework to a clinically relevant classification task: identifying resistance to isoniazid in \mtb{} from whole-genome sequencing data. \mtb{} remains the leading cause of death from a single bacterial pathogen, with over $1.25$ million deaths in $2024$ alone \cite{WHO2024}. Resistance to isoniazid, a first-line anti-tuberculosis drug, arises from spontaneous point mutations rather than horizontal gene transfer, making single nucleotide polymorphism (SNP)-based prediction both feasible and clinically relevant \cite{Waller2023,Nimmo2022}.

\vspace{1em}

Building on this clinical relevance, we apply the experimental frameworks of Belkin et~al.\ \cite{Belkin2019} and Curth et~al.\ \cite{Curth2023} to investigate whether double descent arises in classical machine learning models trained on genomic data from the Comprehensive Resistance Prediction for Tuberculosis (CRyPTIC) consortium \cite{CRyPTIC2022}. Specifically, we train decision trees and gradient boosting regressors to predict isoniazid resistance from whole-genome sequencing data and assess whether prediction error exhibits the characteristic double descent curve. In doing so, we aim to evaluate whether any observed patterns align with the ``unfolding'' hypothesis proposed by Curth et~al.\ \cite{Curth2023}, thereby determining whether double descent constitutes a real generalisation principle or a representational artefact of model parameterisation. We hypothesise that double descent uniquely emerges when model complexity is projected along a unidimensional axis, and that classical bias--variance dynamics reappear when complexity is treated as a multidimensional construct.

\section{Methods}

\subsection*{Data Sources}
\begin{figure}[H]
  \centering
  \includegraphics[width=\linewidth]{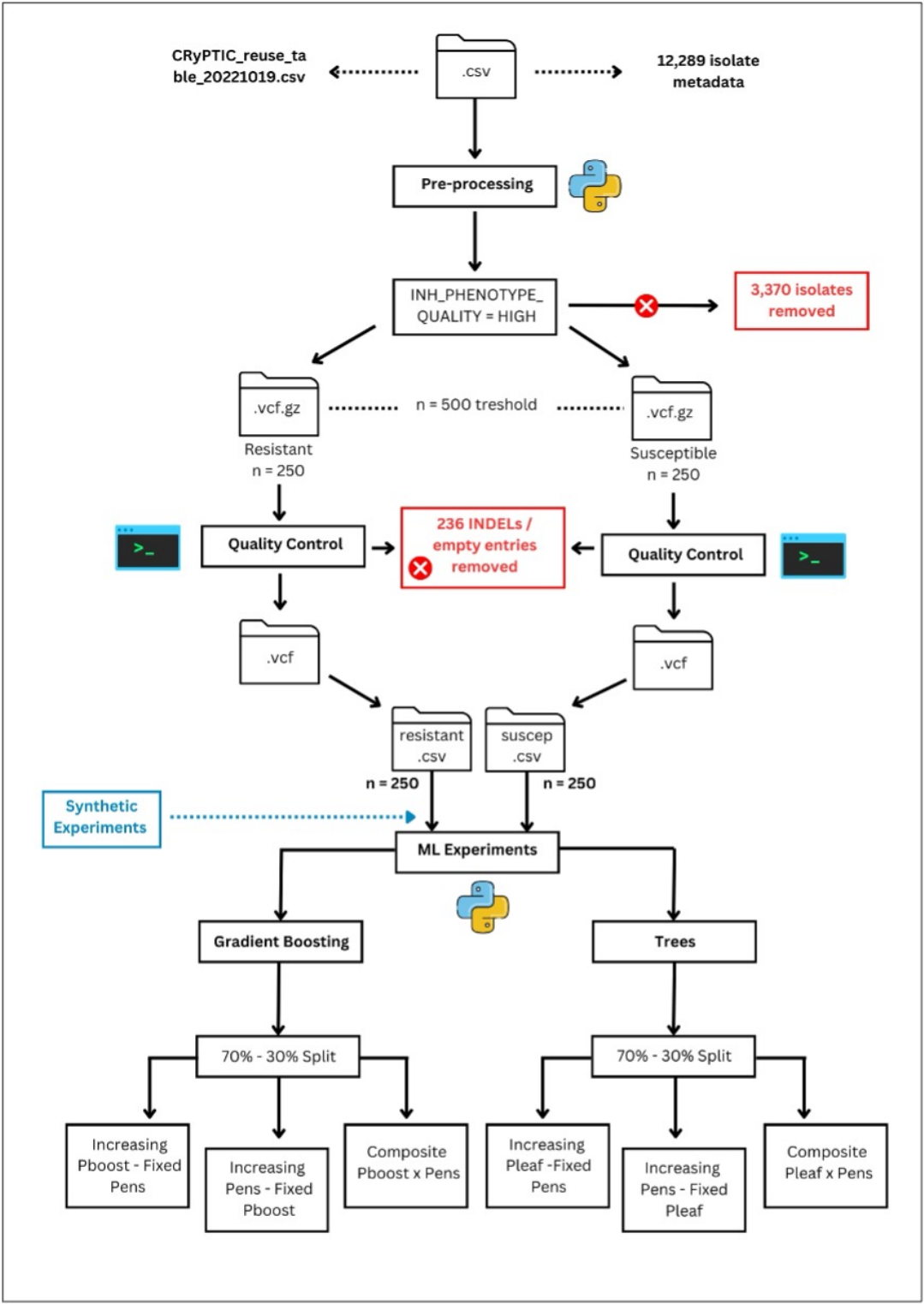}
  \caption{Methodological pipeline. The blue dashed line indicates the branch where synthetic data experiments were conducted, following the same structure as the pipeline used for CRyPTIC data.}
  \label{fig:pipeline}
\end{figure}

Whole-genome sequencing data were sourced from the June 2022 public release of the CRyPTIC consortium, comprising $12{,}289$ \mtb{} isolates from $23$ countries. Each isolate was annotated with phenotypic classifications for resistance or susceptibility to $13$ antibiotics. Associated variant data were obtained in Variant Call Format (VCF), and metadata were retrieved from the accompanying CSV files. Data were accessed from the European Bioinformatics Institute's public FTP repository: \url{https://ftp.ebi.ac.uk/pub/databases/cryptic/release_june2022/reuse/}. An overview of the full data processing and analysis pipeline is presented in Figure~\ref{fig:pipeline}.

\subsection*{Sample Selection and Pre-Processing}
To ensure computational tractability and class balance, we selected a stratified subsample of $n=500$ isolates: $250$ resistant and $250$ susceptible to isoniazid. Only isolates labelled with a ``HIGH'' phenotype quality---defined by CRyPTIC as agreement across at least two minimum inhibitory concentration assays---were retained to reduce label noise. This filtering step removed $3{,}370$ low-confidence samples. Variant data were then parsed from the corresponding VCF files. During quality control, all insertion--deletion mutations (INDELs) and loci with missing genotype calls were removed. This reduced the average number of loci per isolate from $1{,}767$ to $1{,}531$. For each SNP, we extracted four features: genomic position (POS), genotype (GT), read depth (DP), and genotype confidence (GT\_CONF). Genotypes were encoded numerically as $0$ (homozygous reference), $1$ (heterozygous), and $2$ (homozygous alternate). The final feature matrix had dimensions $765{,}413 \times 4$. Although the feature-to-sample ratio was high, no additional dimensionality reduction was applied. This decision follows the conventions of Belkin et~al.\ \cite{Belkin2019} and Curth et~al.\ \cite{Curth2023}, who recommend preserving high-dimensional structure when evaluating double descent, as it facilitates overfitting, thus ensuring a more rapid reach to the interpolation threshold \cite{Ningyuan2022}. 

\subsection*{Machine Learning Framework}
We evaluated three regression-based machine learning models: decision tree regressors, random forest regressors, and gradient boosting regressors. Although the prediction task is inherently binary, we adopted a squared-loss regression framework to align with the methodology of Belkin et~al.\ \cite{Belkin2019} and Curth et~al.\ \cite{Curth2023}, who demonstrated double descent under this loss function. Phenotypic labels were binarised as $y\in\{0,1\}$, and mean squared error ($\MSE$) on the test set was used as the generalisation metric. Model complexity was varied systematically along two orthogonal axes: base learner capacity and ensemble size. For decision tree-based models (including random forests), complexity was parameterised using the number of terminal leaf nodes per tree ($\Pleaf$) and the number of estimators in the ensemble ($\Pens$). Three experimental regimes were implemented. First, we varied $\Pleaf \in \{2,\dots,500\}$ with $\Pens \in \{1,5,10,50\}$ held fixed per sweep. Second, we varied $\Pens \in \{1,\dots,50\}$ with $\Pleaf \in \{20,50,100,500\}$ held fixed per sweep. Third, we used a composite scaling design: $\Pleaf \in \{50,100,200,500\}$ was increased within a single tree up to $L_{\max}$ and then $\Pens$ was scaled from RF1 to RF50, simulating capacity growth past the interpolation threshold \cite{Curth2023}. Gradient boosting models followed the same logic with constraints: base learners had $\Pleaf \le 10$ and we used a high learning rate $\gamma=0.85$ to encourage rapid interpolation \cite{Barbier2025}. In the first experiment, $\Pboost \in \{10,20,50,100,200\}$ with $\Pens \in \{1,5,10,50\}$. In the second, $\Pboost \in \{20,50,100,200\}$ with $\Pens \in \{1,\dots,50\}$. In the third, we fixed $\Pboost=200$ and scaled $\Pens \in \{1,\dots,50\}$. All evaluations used a consistent grid and the same $70\!:\!30$ train--test split. 

\subsection*{Synthetic Baseline}
To validate observed dynamics in a controlled setting, we reproduced all experiments on a synthetic dataset proposed by Friedman (1991) \cite{Friedman1991} and used in contemporary double descent work \cite{Curth2023}. The dataset contained $n=500$ samples and $p=50$ independent features. For each sample, we draw a feature vector $X=(X_1,\dots,X_p)$ with $X_i \stackrel{\text{i.i.d.}}{\sim} U(0,1)$, so that the coordinates are independent and identically distributed on $[0,1]$. The regression target $y$ is generated by the standard Friedman~\#1 formula:
\begin{equation}
  y \;=\; \sin\!\bigl(\pi X_1 X_2\bigr)\;+\;2\,(X_3-0.5)^2\;+\;X_4\;+\;0.5\,X_5\;+\;\varepsilon,
  \qquad \varepsilon \sim \mathcal{N}(0,1).
\end{equation} This benchmark probes non-linear interactions, sparsity (only the first five coordinates are signal-bearing), and additive noise \cite{Belkin2019,Curth2023}. All models used the same hyperparameter grid and a $70\!:\!30$ train--test split.

\subsection*{Code Availability}
All preprocessing, feature extraction, and modelling were conducted using the Cloud Infrastructure for Microbial Bioinformatics \cite{Connor2016}. Variant filtering used Bash scripts; downstream modelling used Python~3.12 with scikit-learn~1.6.1 \cite{Pedregosa2018}, NumPy~2.2.3 \cite{Harris2020}, pandas~2.2.3 \cite{McKinney2010}, Matplotlib~3.10.1 \cite{Hunter2007}, and SciPy~1.15.2 \cite{Virtanen2020}. A fixed random seed ensured reproducibility. All code and documentation: \url{https://github.com/guillermocomesanacimadevila/Demystifying-Double-Descent-in-ML}.

\section{Results and Discussion}

To evaluate whether the double descent phenomenon occurs in classical machine learning models, we trained decision trees and gradient boosting regressors on both real-world genomic data (CRyPTIC) and a synthetic benchmark. Model complexity was varied along two orthogonal axes: learner capacity (e.g., $\Pleaf$ or $\Pboost$) and ensemble size ($\Pens$). This dual-axis framework allowed us to test competing hypotheses: that double descent reflects a generalised learning principle \cite{Belkin2019}, or that it is a projection artefact arising from collapsing separate complexity dimensions \cite{Curth2023}. Across all experiments, our results consistently support the latter.

\subsection*{Composite Complexity Induces Double Descent in Trees and Boosting}
\begin{figure}[H]
  \centering
  \includegraphics[width=0.95\linewidth]{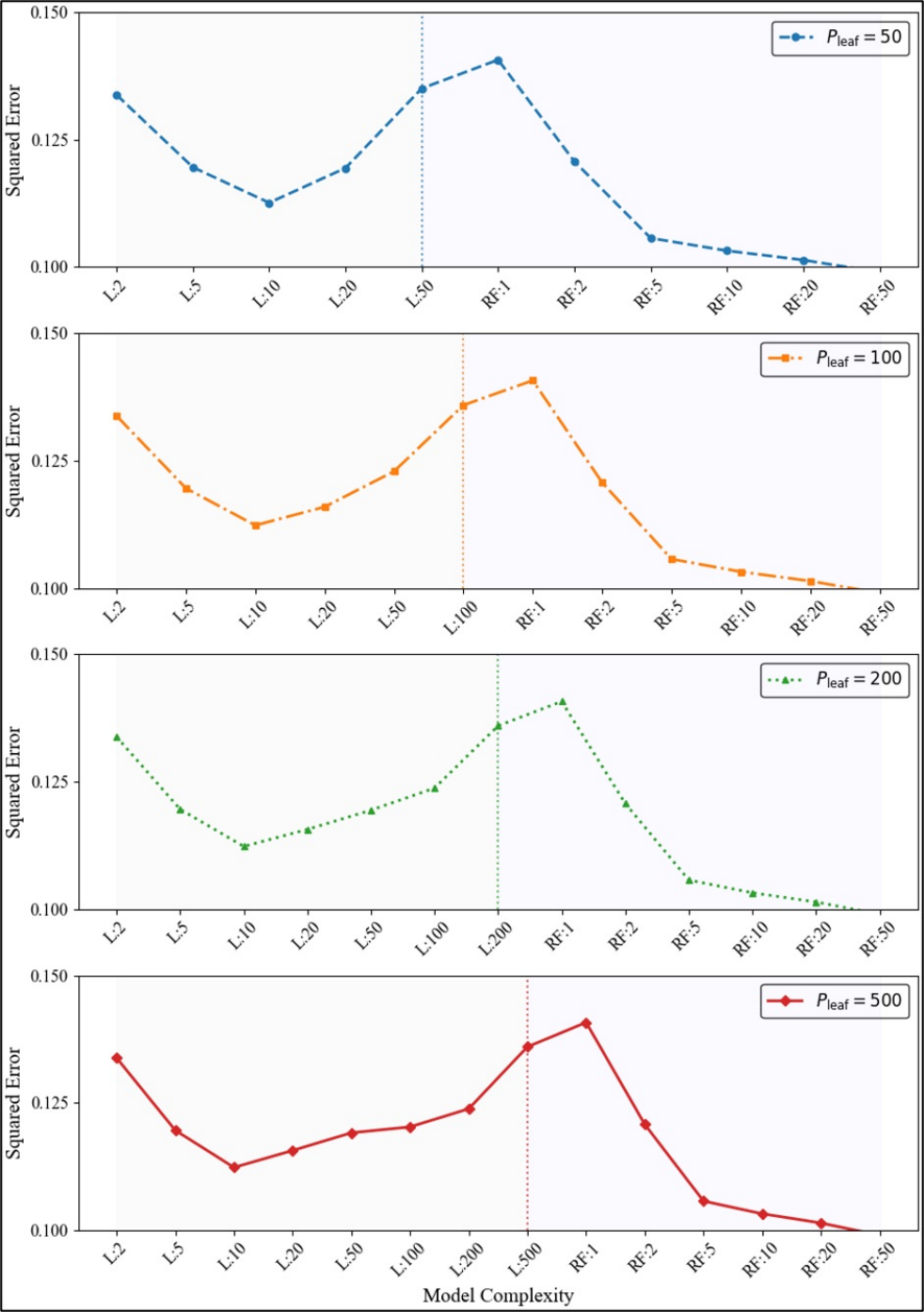}
  \caption{Composite complexity in decision trees and random forests on the CRyPTIC dataset. $\MSE$ is plotted against model complexity for $\Pleaf\in\{50,100,200,500\}$. Within each subplot, complexity increases first by growing single-tree capacity ($L_2$ to $L_{\max}$), then by increasing $\Pens$ (RF1 to RF50). The vertical dotted line marks the interpolation threshold.}
  \label{fig:tree-composite}
\end{figure}

\begin{figure}[H]
  \centering
  \includegraphics[width=1\linewidth]{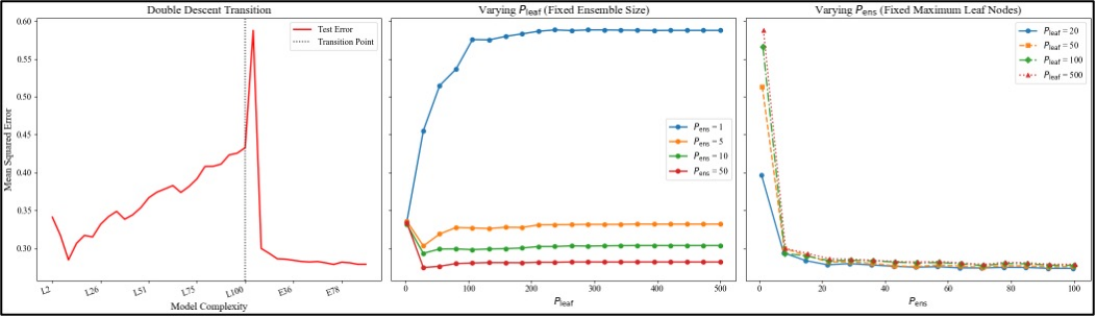}
  \caption{Test $\MSE$ for tree-based models on the synthetic dataset. Left: composite complexity (increasing $\Pleaf$ then $\Pens$). Middle: $\MSE$ vs.\ $\Pleaf$ at fixed $\Pens$. Right: $\MSE$ vs.\ $\Pens$ at fixed $\Pleaf$.}
  \label{fig:synthetic-trees}
\end{figure}

\begin{figure}[H]
  \centering
  \includegraphics[width=1\linewidth]{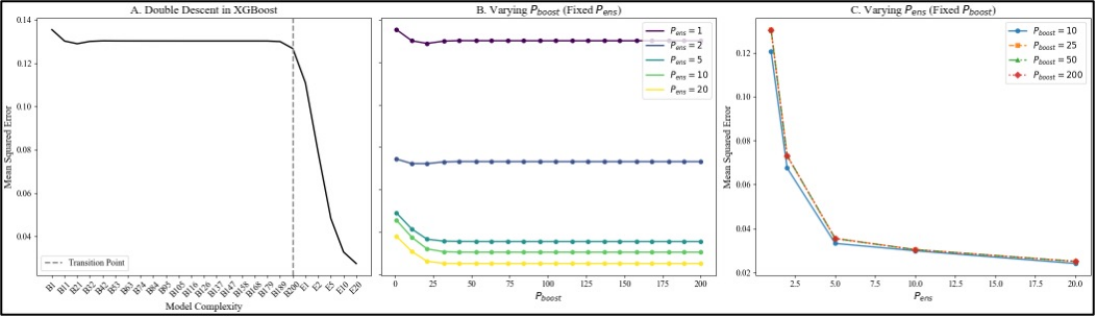}
  \caption{Gradient boosting on the CRyPTIC dataset. (A) Composite complexity: increasing $\Pboost$ then $\Pens$. (B) $\MSE$ vs.\ $\Pboost$ at fixed $\Pens$. (C) $\MSE$ vs.\ $\Pens$ at fixed $\Pboost$.}
  \label{fig:gb-cryptic}
\end{figure}

\begin{figure}[H]
  \centering
  \includegraphics[width=1\linewidth]{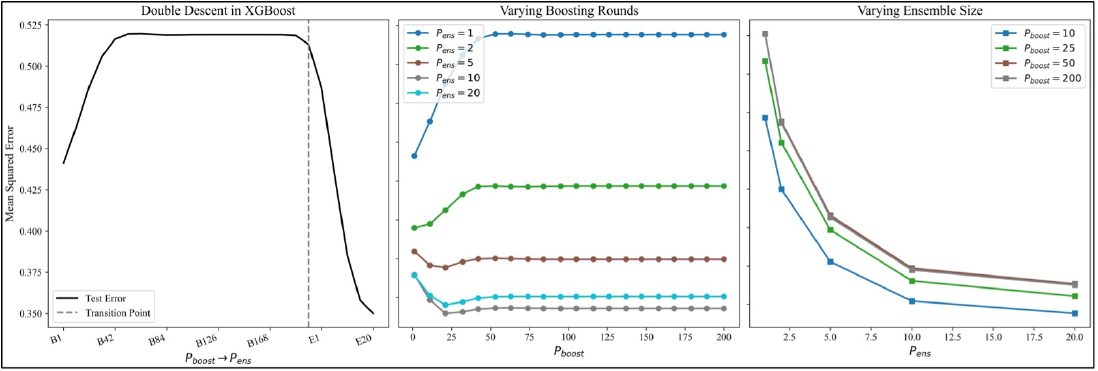}
  \caption{Gradient boosting on the synthetic dataset. (A) Composite complexity: increasing $\Pboost$ then $\Pens$. (B) $\MSE$ vs.\ $\Pboost$ at fixed $\Pens$. (C) $\MSE$ vs.\ $\Pens$ at fixed $\Pboost$.}
  \label{fig:gb-synth}
\end{figure}

When model complexity was increased in a composite manner---first by scaling learner capacity (increasing $\Pleaf$ or $\Pboost$), followed by expanding $\Pens$---a clear double descent pattern emerged in both decision trees and gradient boosting regressors. In decision trees trained on the CRyPTIC dataset (Figure~\ref{fig:tree-composite}), test error initially declined as $\Pleaf$ (in a single tree) increased from $L_2$ to $L_{\max}$, reaching a minimum at $L_{10}$ (e.g., from $0.135$ at $L_2$ to $0.115$ at $L_{10}$ for $\Pleaf=50$). It then rose sharply near the interpolation threshold (marked by the dotted vertical line), peaking at $0.135$--$0.145$ depending on configuration (e.g., $0.140$ at $L_{100}$ for $\Pleaf=100$, $0.140$ at $L_{200}$ for $\Pleaf=200$, and $0.145$ at $L_{500}$ for $\Pleaf=500$). Finally, test error fell again as $\Pens$ increased from RF1 to RF50, reaching values as low as $0.100$--$0.103$ across all settings. This non-monotonic behaviour was observed consistently across all four $\Pleaf$ configurations. While the position and height of the error peak varied slightly, each curve exhibited the hallmark shape of double descent. The same trajectory was observed in the synthetic dataset (Figure~\ref{fig:synthetic-trees}, left). Gradient boosting models demonstrated similar dynamics under composite scaling. On the CRyPTIC dataset (Figure~\ref{fig:gb-cryptic}A), $\MSE$ declined from $0.118$ at $\Pboost=10$ to a minimum of $0.081$ at $\Pboost=200$, then rose near the interpolation threshold, before falling again to $0.074$ at $\Pens=50$. The synthetic data mirrored this behaviour (Figure~\ref{fig:gb-synth}A), with $\MSE$ decreasing from $0.099$ to $0.063$, peaking at $0.121$, and then falling again to $0.059$.

\vspace{1em}

\vspace{1em}

\clearpage

\subsection*{Axis-Specific Scaling Reveals Bias--Variance Trade-off}
\begin{figure}[H]
  \centering
  \includegraphics[width=1\linewidth]{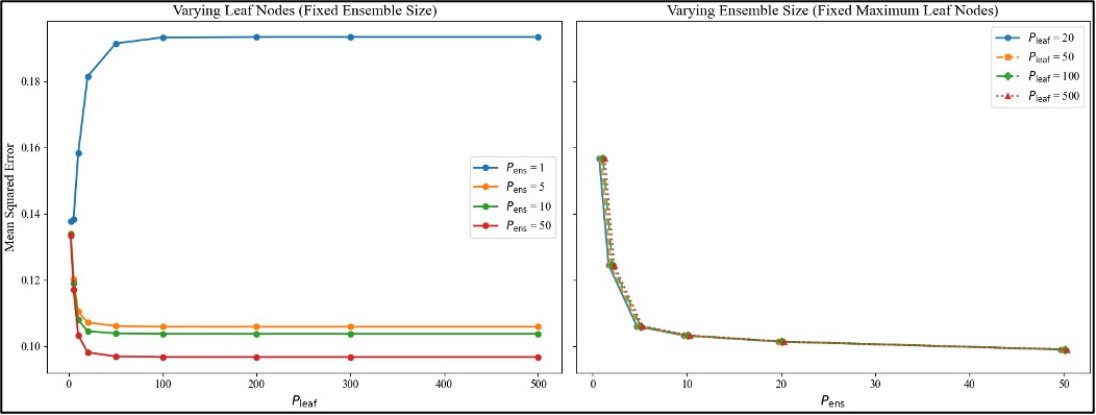}
  \caption{Independent sweeps on CRyPTIC. Left: $\MSE$ vs.\ $\Pleaf$ at fixed $\Pens$. Right: $\MSE$ vs.\ $\Pens$ at fixed $\Pleaf$.}
  \label{fig:axis-sweeps}
\end{figure}

When model complexity was varied along a single axis---either by increasing learner capacity or ensemble size independently---the double descent pattern disappeared. Instead, generalisation behaviour aligned with the classical bias--variance trade-off. In decision trees trained on CRyPTIC (Figure~\ref{fig:axis-sweeps}), increasing $\Pleaf$ at fixed $\Pens$ resulted in a characteristic U-shaped curve. For example, with $\Pens=1$, $\MSE$ decreased from $0.137$ at $\Pleaf=2$ to a minimum of $0.107$ at $\Pleaf=20$, but rose sharply to $0.194$ by $\Pleaf=500$. This sharp increase highlights overfitting in high-capacity learners without variance control, as described by \cite{RocksMehta2022}. In contrast, holding $\Pleaf$ constant and increasing $\Pens$ reduced $\MSE$ smoothly and monotonically, producing an L-shaped curve. For example, at $\Pleaf=100$, test error dropped from $0.136$ at $\Pens=1$ to $0.097$ at $\Pens=50$. These trends were replicated in the synthetic dataset (Figure~\ref{fig:synthetic-trees}).

\vspace{1em}

Gradient boosting models showed an analogous pattern under axis-specific scaling (Figures~\ref{fig:gb-cryptic}B--C; \ref{fig:gb-synth}B--C). At low ensemble sizes, increasing $\Pboost$ introduced overfitting. In CRyPTIC (Figure~\ref{fig:gb-cryptic}B), test $\MSE$ rose from $0.123$ at $\Pboost=10$ to $0.134$ at $\Pboost=200$ with $\Pens=1$, reflecting the high-variance behaviour of overparameterised learners \cite{James2021}. A similar trend appeared in the synthetic dataset (Figure~\ref{fig:gb-synth}B), where $\MSE$ increased from $0.094$ to $0.147$ across the same boosting range. Conversely, increasing $\Pens$ while keeping $\Pboost$ fixed consistently reduced test error. At $\Pboost=50$, $\MSE$ on CRyPTIC (Figure~\ref{fig:gb-cryptic}C) fell from $0.108$ at $\Pens=1$ to $0.042$ at $\Pens=20$. The synthetic dataset (Figure~\ref{fig:gb-synth}C) mirrored this L-shaped descent, with $\MSE$ dropping from $0.099$ to $0.048$.  Across all models, one axis---$\Pleaf$ in trees or $\Pboost$ in boosting---exerted a disproportionate influence on test error, while the other axis ($\Pens$) tended to improve performance or, at worst, leave it unchanged. This asymmetry highlights a consistent “bigger is better” effect with respect to $\Pens$, contrasting with the more volatile behaviour observed when scaling learner capacity alone.

\vspace{1em}

\clearpage

\subsection*{Comparison with Existing Literature}
While our error curves reflect the double descent dynamics described by Belkin et al. \cite{Belkin2019}, they align more closely with the ``unfolding'' hypothesis of Curth et al. \cite{Curth2023}. Rather than viewing double descent as a universal feature, the unfolding framework suggests that the phenomenon arises when distinct generalisation behaviours---underfitting, interpolation, and overparameterisation---are projected onto a single axis of model complexity. Our results support this view: when capacity and ensemble size are disentangled, the apparent double descent resolves into more interpretable U- and L-shaped curves. This perspective also challenges common assumptions about the robustness of ensemble methods. Random forests and gradient boosting are often viewed as resistant to overfitting, largely due to variance-reducing techniques like averaging in ensembles and regularisation strategies such as shrinkage and subsampling \cite{Schonlau2020,Park2020}. 

\vspace{1em}

However, our findings suggest that this robustness is conditional on how model complexity is scaled. When learner capacity is increased---for example, by upregulating $\Pleaf$---without a corresponding increase in $\Pens$, test error can rise sharply---an effect often overlooked in standard hyperparameter tuning workflows, which typically vary only one parameter at a time \cite{Barbier2025,Raschka2018}. This interpretation helps reconcile conflicting findings in the literature. For example, \cite{BuschjaegerMorik2022} observed no double descent in well-tuned random forests---that is, models tuned along a single complexity axis. Our results confirm that under axis-specific tuning, test error behaves predictably. However, when complexity is scaled sequentially across both axes---as in our composite regime---the double descent curve reliably re-emerges \cite{Schaeffer2023,Curth2023}. These findings suggest that double descent is not a property of algorithm type, but of the trajectory through complexity space during training.

\subsection*{Practical Implications for Model Tuning}
Our findings carry important implications for model selection and tuning. They reaffirm the continued relevance of classical bias--variance theory \cite{James2021,Domingos2000}, but only when model complexity is treated as a multidimensional concept. As highlighted by \cite{Curth2023,Schaeffer2023}, the apparent breakdown of generalisation theory in overparameterised models often stems not from a failure of the theory itself, but from conflating multiple complexity axes into one. The error peak near the interpolation threshold---central to the double descent narrative \cite{Belkin2019}---is better understood as a misalignment between capacity and variance control. In practice, this means that sharp increases in test error may not reflect flaws in the model or data noise but can instead arise from how hyperparameters are scaled during training. For example, increasing $\Pleaf$ or $\Pboost$ without adjusting $\Pens$ can push the model into a high-variance regime. These instability points---observed in both our work and previous studies \cite{Barbier2025,BuschjaegerMorik2022,Curth2023}---are frequently misinterpreted as poor model performance, when they are actually artefacts of composite scaling. Therefore, our results support the unfolding hypothesis. Moreover, as \cite{Schaeffer2023} observe, tuning multiple hyperparameters simultaneously---such as $\Pleaf$ and $\Pens$---can obscure which one is driving performance changes. By varying them independently, practitioners can disentangle their effects, diagnose variance-related instabilities, and more effectively tune model behaviour. This targeted approach not only improves the clarity of generalisation patterns but also guides more efficient and reliable hyperparameter tuning \cite{Schaeffer2023,Curth2023}.

\subsection*{Strengths and Limitations}
This study is, to our knowledge, the first to systematically evaluate the double descent phenomenon in classical machine learning models applied to real-world biological data. By applying the unfolding hypothesis to decision trees and gradient boosting regressors across both synthetic and clinical datasets, we provide empirical support for a multidimensional view of generalisation. Our findings extend the composite scaling framework introduced by \cite{Curth2023} and highlight the value of axis-aware tuning in practice. Several limitations, however, warrant discussion. First, our analysis focuses exclusively on tree-based models, which may limit the generalisability of results to other algorithmic families, such as support vector machines, where the dynamics of double descent have not yet been critically examined \cite{BuschjaegerMorik2022}. Second, the scope of our study is restricted to classical (non-deep) learners. Whether the unfolding hypothesis, as formulated by \cite{Curth2023}, offers a valid or useful framework for understanding double descent in deep neural networks remains an open question. Lastly, we intentionally avoided dimensionality reduction to align with prior work on double descent (e.g., \cite{Belkin2019,Curth2023}). However, biological data is often inherently noisy \cite{Li2016}, so this choice may have further amplified variance by retaining irrelevant or redundant features \cite{ChiziMaimon2009}. For example, SNPs unrelated to isoniazid resistance are likely irrelevant, while redundancy may arise from co-inherited variants within genes like \textit{katG}, which tend to be in linkage disequilibrium due to the low recombination rate in \mtb{} \cite{Marney2018}.

\section{Conclusion}
This study investigated the double descent phenomenon in classical machine learning models---specifically decision trees and gradient boosting regressors---applied to both synthetic data and a clinically relevant genomic prediction task. By independently and jointly scaling model complexity along two orthogonal axes---learner capacity ($\Pleaf$, $\Pboost$) and ensemble size ($\Pens$)---we showed that double descent emerges consistently under composite scaling, but not when these axes are varied in isolation. These results support the unfolding hypothesis \cite{Curth2023}, which argues that double descent arises from conflating distinct generalisation regimes onto a single complexity axis. Contrary to the notion that ensemble methods are inherently resistant to overfitting, our findings demonstrate that gradient boosting and random forests can exhibit double descent when model capacity is increased without adequate variance control. Across both datasets, $\Pens$ consistently acted as a stabilising factor, revealing its role as an implicit regulariser. This highlights the importance of understanding not just the magnitude of model complexity, but how it is structured and scaled. While our focus was limited to tree-based models, the experimental design introduced here offers a general framework for testing double descent in other learning algorithms. Future work should extend this framework to support vector machines and neural networks and explore how factors such as dimensionality reduction, feature redundancy, and label noise shape generalisation dynamics in high-capacity regimes. Overall, our findings reinforce the continued relevance of classical bias--variance theory---provided model complexity is treated as a multidimensional construct.

\clearpage


\end{document}